\documentclass[10pt,twocolumn,letterpaper]{article}

\usepackage[accsupp]{axessibility}  

\makeatother\usepackage{cite}
\usepackage{amsmath,amssymb,amsfonts}
\usepackage{graphicx}
\usepackage{textcomp}
\usepackage{xcolor}
\usepackage{comment}
\usepackage{booktabs}
\usepackage{tikz}
\usepackage{caption}

\usetikzlibrary{positioning}
\tikzstyle{Mytext} = [text width=0.1*\linewidth]

\newcommand{\ad}[2]{{\color{lightgray}#1}{\color{green}#2}}

\usepackage[pagebackref,breaklinks,colorlinks]{hyperref}

\usepackage[capitalize]{cleveref}
\crefname{section}{Sec.}{Secs.}
\Crefname{section}{Section}{Sections}
\Crefname{table}{Table}{Tables}
\crefname{table}{Tab.}{Tabs.}

\def\BibTeX{{\rm B\kern-.05em{\sc i\kern-.025em b}\kern-.08em
    T\kern-.1667em\lower.7ex\hbox{E}\kern-.125emX}}

\newcommand{\gf}[2]{{\color{lightgray}#1}{\color{purple}#2}}
\begin{document}

\title{On the Importance of Large Objects in CNN Based Object Detection Algorithms}

\author{
Ahmed Ben Saad$^{1,2}$~~~Gabriele Facciolo$^{1}$~~~Axel Davy$^{1}$\vspace{0.2cm}\\
$^{1}$Université Paris-Saclay, CNRS, ENS Paris-Saclay, Centre Borelli, 
91190, Gif-sur-Yvette, France~~~~\\
$^{2}$Schlumberger AI Lab
}

\maketitle

\begin{abstract}
Object detection models, a prominent class of machine learning algorithms, aim to identify and precisely locate objects in images or videos. However, this task might yield uneven performances sometimes caused by the objects sizes and the quality of the images  and labels used for training.
In this paper, we highlight the importance of large objects in learning features that are critical for all sizes. Given these findings, we propose to introduce a weighting term into the training loss. This term is a function of the object area size. We show that giving more weight to large objects leads to improved detection scores across all object sizes and so an overall improvement in Object Detectors performances (+2~p.p. of mAP on small objects, +2~p.p. on medium and +4~p.p. on large on COCO val 2017 with InternImage-T). Additional experiments and ablation studies with different models and on a different dataset further confirm the robustness of our findings.
\end{abstract}

\section{Introduction}
Object detection is a fundamental task in computer vision with applications in a variety of fields (autonomous vehicles, surveillance, robotics, ...). It has been studied in computer vision since the dawn of automatic Image Processing~\cite{viola2001rapid,dalal2005histograms,khan2012color}. The surge of Convolutional Neural Networks (CNNs)~\cite{krizhevsky2017imagenet} has revolutionized the field leading to a proliferation of methods~\cite{redmon2016look,lin2017feature,he2017mask,tan2020efficientdet,wang2023internimage} and substantial improvements in detection scores.
\begin{figure}[t]
    \centering
    \includegraphics[trim={20 0 10 0},clip, width=\columnwidth, height=.95\linewidth]{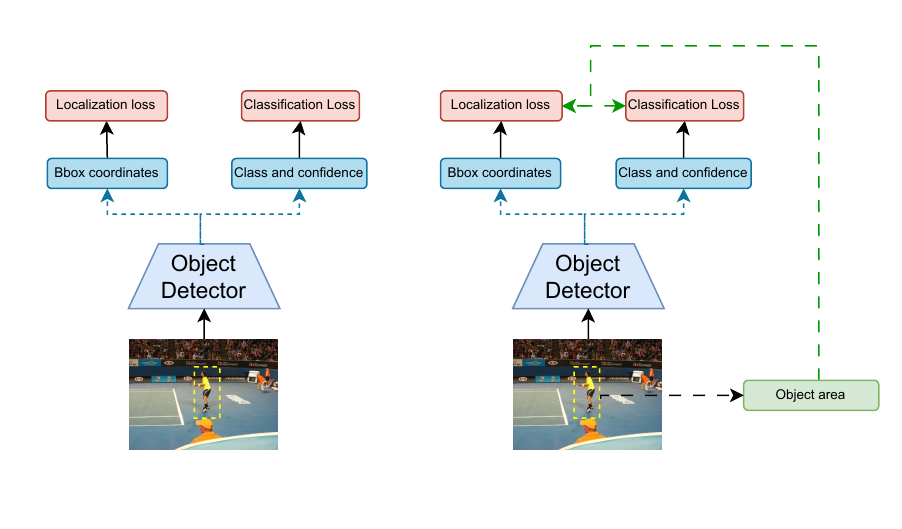}
\vspace{-4em}
    \caption{Overview of the proposed weighting policy (right) compared to a general object detection framework (left). The area of each bounding boxes is computed (black dotted arrow) then its log is taken as a sample weight for the corresponding bounding box in both classification and localization losses (green dotted arrows). This gives more importance to large objects and performances across all sizes benefit from it.}
    \label{fig:cocosizes}
\end{figure}

Researchers have proposed several variants of object detection models, including one-stage~\cite{liu2016ssd, redmon2016look, lin2017focal,tian2019fcos} and two-stage detectors~\cite{girshick2015fast, he2017mask, lin2017feature, pramanik2021granulated}, to improve the speed and accuracy of object detection. Furthermore, novel techniques, such as attention mechanisms~\cite{vaswani2017attention,carion2020end, liu2021swin} and anchor-free object detection~\cite{tian2019fcos, kong2020foveabox,cheng2022anchor}, have emerged to further improve the performances of existing models. In this paper, we aim to focus on object detection models and analyze their underlying mechanisms for locating objects within an image.

Detection datasets usually contain a large number of easy examples and a small number of hard ones. Automatic selection of these hard examples can make training more effective and efficient~\cite{Shrivastava_2016_CVPR}. Different data sampling techniques were proposed depending on the criterion for selecting the hard samples during training. These criteria include high current training loss~\cite{DBLP:journals/corr/Simo-SerraTFKM14}, Foreground/Background ratio unbalance~\cite{7485869,lin2017focal},  IoU-unbalance shifting towards hard examples~\cite{DBLP:journals/corr/abs-1904-02701} and class unbalance~\cite{Pang_2020_CVPR}.

The influence  on detection performance of object size distribution of a training dataset is less examined subject in the literature.
Common wisdom would dictate that if the final goal is to have a maximum performance for a given size of objects - say small objects - more emphasis during training should be given to these target objects. Our work shows that reality can be counter-intuitive, as we find that giving more focus to large objects can improve performance for all object sizes, including small ones.
Indeed, we find that a simple change in the training loss can increase performance for various object detectors.
The loss functions of object detection can be categorized as two sorts: the classification loss and the localization loss. The first is used to train a classification head that will detect and, in the case of multiclass object detection, categorize the target object. The second is used to train a head that will regress a rectangular box to find the target object.
We propose to incorporate the sample weight function in the total loss computation, including the classification term (see Figure~\ref{fig:cocosizes}). By assigning less weight to smaller objects and more weight to larger objects, the model learns effectively from both small and large objects.

Through empirical evaluations and ablations, we validate the effectiveness of the proposed weight function and demonstrate its potential for advancing the state-of-the-art in object detection.
Our contribution are the following:
\begin{itemize}
  \item We verify that learning on large objects leads to better detection performance than learning on small objects.
  \item We propose a simple loss re-weighting scheme that gives more emphasis to large objects, which results in an overall improvement of object detectors' performances across all objects sizes.
  \item We analyze for which object detection sub-tasks the performance gains are most seen, improving the understanding of the impact of the loss re-weighting.
\end{itemize}

\section{Related work}
\label{sec:related_work}

Besides the use of geometrical data augmentation techniques, over the years object detector architectures have incorporated more and more elements to improve performance across object scales.
In this section,  we review some of the models that we deem important for their influence or performances. Mainly highlighting the proposed ideas to deal with the different object sizes. We then focus on data augmentation, how it has been used for the same goal and its limitations.

\paragraph{Feature Pyramid Networks (FPN)}
Feature Pyramid Networks (FPN) is a widely used module proposed by Lin et al.~\cite{lin2017feature} that addresses the limitations due to having one common prediction output for all object scales. 
More specifically it proposed to extract features at different levels of a backbone convolutional network~\cite{redmon2013darknet,tan2020efficientdet,lin2017feature}, and merge them back in an inverted feature pyramid. Then each level of the inverted feature pyramid has a dedicated detection branch dedicated to objects of a given size range. The performance gains can be attributed to capturing semantic information at higher resolutions while maintaining spatial information at lower resolutions.

\begin{figure}[t]
\centering
\resizebox{\columnwidth}{0.4\linewidth}{%
\begin{tabular}{lll}
\includegraphics[height= \linewidth ]{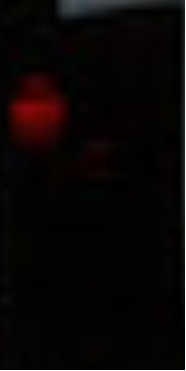} & \includegraphics[height= 2\linewidth ]{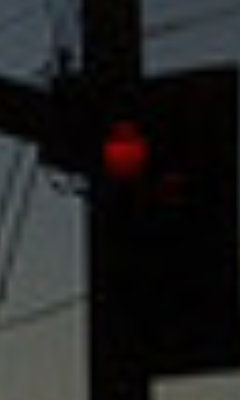} & \includegraphics[height=3 \linewidth ]{figs/small/1/big.png} \\
\includegraphics[height= \linewidth ]{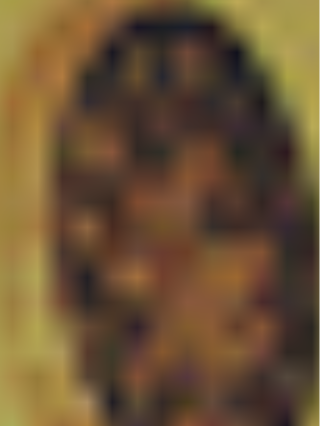} & \includegraphics[height=2 \linewidth ]{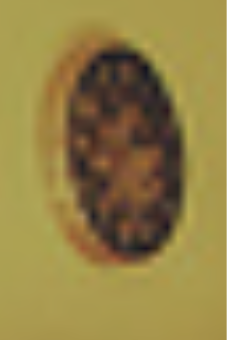} & \includegraphics[height=3 \linewidth ]{figs/small/2/big.png} \\

\includegraphics[height= \linewidth ]{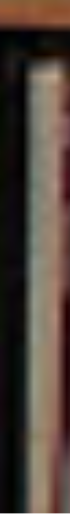} & \includegraphics[height=2 \linewidth ]{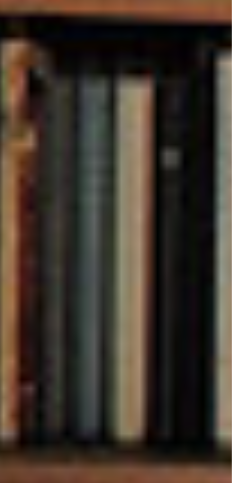} & \includegraphics[height=3 \linewidth, width = 4.495 \linewidth ]{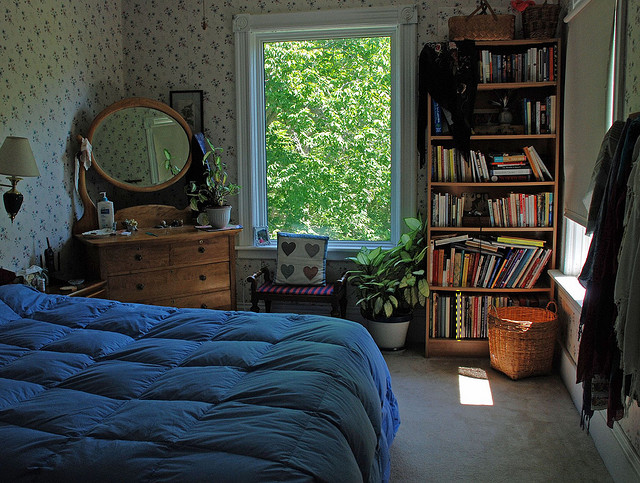} \\

\end{tabular}%
}
\caption{Example of some small objects cropped without or with small context (first and second columns) and their entire context in the image (third column, the objects are highlighted in yellow bounding boxes, zoom for better view). We focus on a traffic light in the first row, a clock in the second and a book in the third. We see from these examples that in the case of small objects, it is difficult, even to a human eye, to correctly label the designated object. Also, the more context we have, the easier is the classification. This also applies to CNNs.}
\label{fig:ctx}
\end{figure}

\paragraph{YOLO}
YOLO (You Only Look Once), proposed by Redmon et al.~\cite{redmon2016look}, is a real-time anchor based one-stage object detection system that 
uses a single neural network to simultaneously predict object bounding boxes and class probabilities in real time and directly from input images. Achieving state-of-the-art speed and accuracy. 
Since its inception, YOLO has undergone several evolutions to enhance its performance.
YOLOv2~\cite{Redmon_2017_CVPR} improved upon the original architecture by introducing anchor boxes to enable the model to efficiently detect objects of different aspect ratios and sizes.
YOLOv3~\cite{DBLP:journals/corr/abs-1804-02767} incorporated a feature pyramid network, which enabled the model to effectively capture objects at multiple scales.
 YOLOv4~\cite{DBLP:journals/corr/abs-2004-10934} adopted the CSPDarknet53~\cite{wang2020cspnet} backbone, which improved the model's capacity to extract complex features. It also incorporated the PANet~\cite{wang2019panet} module, which performed feature aggregation across different levels of the network,further improving object detection at various scales. 
YOLOv5~\cite{jocher2021ultralytics} is a PyTorch implementation of YOLO, characterized by practical quality-of-life improvements for training and inference. In terms of performance it is comparable to YOLOv4.

\paragraph{TTFNet}
TTFNet~\cite{liu2020training} is a derivative of CenterNet~\cite{zhou2019objects}, which defines an object as a single point (the center point of its bounding box). It uses keypoint estimation to find center points and regresses to all other object properties. TTFNet speeds-up the training of CenterNet by predicting bounding boxes not only at the center pixel but also around it using a Gaussian penalization.  Several weighting schemes were considered and the authors found that the best performance was reached by normalizing the weights then multiplying by the logarithm of the area of the box. The localization loss is then normalized by the sum of the weights present in the batch. Inspired by this approach we propose to add the logarithmic weighting also to the other terms, namely localization and classification. Other works such as FCOS~\cite{tian2019fcos}, have studied the impact of the bounding boxes areas on the training, but to the best of our knowledge, none have proposed a weighting scheme to focus on large objects. In FCOS all the pixels of the bounding box contribute to its prediction, but the subsequent loss is averaged among all pixels. Its implementation was later extended as FCOS Plus,\footnote{\url{https://github.com/yqyao/FCOS_PLUS}} which reduces the learning region to a center region inside the box.

\paragraph{DETR}
DETR (Detection Transformer)~\cite{carion2020end} introduces a transformer-based architecture to object detection that enables simultaneous prediction of object classes and their bounding box coordinates in a single pass. Notably, DETR utilizes a global loss function based on sets, allowing it to effectively handle variable object counts through the integration of self-attention mechanisms and positional encodings.

\paragraph{InternImage}
InternImage, proposed by Wang et al.~\cite{wang2023internimage}, is a large-scale CNN-based foundation model which capitalizes on increasing the number of parameters and training data, similar to Vision Transformers~\cite{DBLP:journals/corr/abs-2010-11929}.  InternImage employs deformable convolutions~\cite{DBLP:journals/corr/DaiQXLZHW17} as its core operator, allowing it to capture richer contexts in object representations. 
Moreover, InternImage incorporates adaptive spatial aggregation conditioned by input and task information, reducing the strict inductive bias commonly observed in traditional CNNs. InternImage has attained improved object detection results and currently holds high ranks in evaluation scores across different datasets. As we will see we can further improve the performance of InternImage by training with a size-dependent weighting term.

\paragraph{Data Augmentation}
Data augmentation is a powerful solution to enhance the performance of object detection models across all object sizes~\cite{kaur2021data,shorten2019survey}. By applying transformations to the training dataset, data augmentation techniques introduce diversity and expand the representation of objects at different scales. Augmentations such as random scaling, flipping, rotation, and translation enable the models to learn robust features for accurate detection of both small and large objects. Augmentations specifically designed for small objects, such as random patch copy-pasting and pixel-level augmentations~\cite{DBLP:journals/corr/abs-1902-07296}, help alleviate issues related to low-resolution details and limited contextual information. Similarly, augmentations that preserve spatial context and prevent information loss during resizing or cropping~\cite{maharana2022review,xu2023comprehensive} assist in handling large objects.
However, it is important to note that data augmentation techniques have limitations when it comes to object sizes. While augmentations can introduce diversity and expand the representation of objects, upscaling objects does not inherently yield additional information. Increasing the size of small objects through augmentation may improve their visibility, but it does not provide additional contextual details or features that were not present in the original image. On the other hand, downscaling or resizing larger objects can potentially lead to the loss of important information and fine-grained details, which may hinder accurate detection.

Little attention has been given to the content of the dataset itself (with exception of annotation errors~\cite{benenson2019large, hoiem2012diagnosing}), in particular the impact of object size distribution on the detection performance across all scales. 
In the next section, we highlight the importance of features learned from large objects on the overall performances of object detectors.

\section{On the importance of objects sizes}
%

\begin{table}
\centering
\resizebox{0.75\columnwidth}{!}{
\begin{tabular}{|l|l|l|l|}
\hline
       & Range           & ratio$_{train}$ & ratio$_{val}$      \\ \hline
Small  & [$0$, $32^2$]  & 0.27 & 0.28        \\ 
Medium & [$32^2$ , $96^2$]   & 0.45 & 0.44  \\ 
Large  & [$96^2 $ , $+\infty$[  & 0.27& 0.29 \\ \hline
\end{tabular}%
}

\caption{Objects sizes ranges in COCO dataset.}
\label{tab:sizes}
\end{table}



\begin{table*}
\centering
\captionsetup{justification=centering}
\resizebox{1.7\columnwidth}{!}{%
\begin{tabular}{|c|cc|cc|cc|cc|}
\hline
                       & \multicolumn{2}{c|}{Small}      & \multicolumn{2}{c|}{Medium}     & \multicolumn{2}{c|}{Large}      & \multicolumn{2}{c|}{All}        \\ 
Scores                        & mAP   & mAR   & mAP   & mAR   & mAP   & mAR   & mAP   & mAR   \\\hline

Only pretrain on Small+Medium & 0.265 & 0.425 & 0.451 & 0.676 & 0.401 & 0.628 & 0.281 & 0.512 \\
Only pretrain on Large & 0.187 & 0.356 & 0.325 & 0.605 & 0.482 & 0.731 & 0.255 & 0.426 \\
Pretrain on Small+Medium then finetune on all                  & 0.253 & 0.413 & 0.424 & 0.667 & 0.446 & 0.695 & 0.296 & 0.521 \\
Pretrain on Large then finetune on all                         & 0.271 & 0.433 & 0.487 & 0.681 & 0.542 & 0.753 & 0.350 & 0.604 \\
Reference scores (Train directly on the entire dataset) & \textbf{0.297} & \textbf{0.456} & \textbf{0.540} & \textbf{0.718} & \textbf{0.674} & \textbf{0.833} & \textbf{0.372} & \textbf{0.660} \\ \hline
\end{tabular}%
}

\caption{Test mAP and mAR scores for different pretraining object sizes on COCO val 2017. The finetuning step (if it exists) is done while freezing the encoder part of the network. Note that pretraining on large objects improves the scores for all sizes compared to pretraining on small/medium objects}
\label{tab:pretrain_test}
\end{table*}

Datasets such as COCO incorporate a diverse set of objects of various sizes. However, detecting large objects presents different challenges compared to small ones. Large objects have rich details and texture, which might have to be interpreted or ignored, but this rich information are usually enough to know what they are without surrounding context. Small objects differ in that the surrounding context has significant importance in their interpretation.  As an illustration of this fact, Figure~\ref{fig:ctx} shows a set of cropped small objects without or with their context.
We tend to imagine that small object detection depends mainly on the earlier stages of a backbone. However, this observation implies that the latest stages of the backbone have features that capture large objects, but also the context needed to detect small ones. 
As a result all object sizes need good quality features at all levels of the network backbone. The intuition behind our research is that having a variety of object sizes helps learn high quality features at all sizes, and that emphasizing the importance of large objects in the loss is even better.

This intuition can be verified by the following experiment: Given an object detector (YOLO v5~\cite{jocher2022ultralytics} in this case) and a training dataset (COCO~\cite{cocodataset}) we start by initializing the model with random weights and pretrain it using only large objects. We used the size ranges used by the authors of YOLO v5 in their github repository\footnote{\url{https://github.com/ultralytics/yolov5}} and shown in Table~\ref{tab:sizes}. We then freeze the encoder layers and fine-tune the model on all the training data. We also repeat the same procedure but using the small and medium data for pretraining. The results of train and test mAP and mAR are shown in Table~\ref{tab:pretrain_test}. The goal of these experiments is to observe the quality of the learned backbone features for various object sizes when trained exclusively on large or small+medium objects.

We can see that, despite the relatively lower quantity of large objects compared to the rest of the dataset, the model pretrained on large objects and finetuned on the entire dataset performs better across all sizes.
This means that features for bigger objects are more generic and can be used to detect at all object sizes including smaller ones. This is less the case for features learned on small objects.

Another interesting point is that the network trained only on small and medium objects performs worse on these objects than the network trained on the whole dataset. In fact even the network using the backbone pretrained only on large objects and finetuned on the entire dataset has better detection performance on small objects. This highlights the argument that large objects help learn more meaningful features for all scales. 


\section{Proposed method}
\subsection{Weight term}
To effectively leverage the large-sized objects for enhancing model performance, we propose the inclusion of a weight term in the loss functions specifically designed for object detection tasks
\begin{equation}
W_{i} = log(h_i \times w_i),
\label{eq:log_weights}
\end{equation}
where $h_i$ is the i-th object height and $w_i$ is its width. 

For example, let us consider the YOLO v5 loss
\begin{equation}
   L_{total} =  \lambda_1 L_{classif} + \underbrace{ L_{confidence} + \lambda_2 L_{CIoU}}_{L_{detection}}. 
     \label{eq:loss}
\end{equation}
At each training step, the loss is calculated as an average over all batch samples
\begin{equation}
  L_{a,batch} = \frac{1}{N_b} \sum_{i \in B_{batch}}L_{\psi}(i,\Hat{i})  
  \label{eq:lossbatch}
\end{equation}
with $\psi \in \{confidence, classif, CIoU \}$, $N_b$ the number of bounding boxes in the batch, $B_{batch}$ the set of the bounding boxes in a batch, $i$ the prediction over one bounding box and $\Hat{i}$ the corresponding ground truth. We modify $L_{\psi,batch}$ to incorporate the weights $W_i$ with
\begin{equation}
    L_{\psi,batch}^{new} = \sum_{i \in B_{batch}} \alpha_i L_\psi(i,\Hat{i}),
    \label{eq:log_weighting_scheme}
\end{equation}
where $\alpha_i = \frac{W_i}{\sum_{k=1}^{N_b}W_k}$.
This term aims to assign higher weights to larger objects during the training  and thus encourages the models to focus more on learning from them. On the other side, small objects get a reduced impact on learning, as the sum of the weights in the batch is normalized. Yet, the slow increase of the logarithm means that no object size is negligible in the loss.

As mentioned in Section~\ref{sec:related_work}, the weighting term~\eqref{eq:log_weighting_scheme} was already used in TTFNet. However, contrary to TTFNet, which incorporated this weight in its size regression loss (GIoU), we use it on both the localization and classification loss terms.
We justify this choice by an ablation study in Section~\ref{sub:ab}.

The inclusion of the weight term in the loss functions encourages the models to prioritize the accurate detection and localization of large objects. This leads to more discriminative features and better contextual understanding, particularly concerning larger objects. And as a consequence, the models become better equipped to handle also small objects.

Furthermore, the weight term helps to address the inherent dataset bias towards smaller objects by explicitly giving larger objects more prominence during training. This bias correction allows the models to learn more effectively from the limited number of large objects present in the dataset, bridging the performance gap between small and large object recognition. For example, in Table~\ref{tab:bars}, the ratio of each object size
\begin{equation}
  r_{size} = \frac{\#Objects_{size}}{\#Objects}  
\end{equation}
is compared to the weighted sum of these objects
\begin{equation}
    r'_{size} = \frac{\sum_{object \in size} log(w_{object}h_{object})}{\sum_{Objects} log(w_{object}h_{object})}
\end{equation} on COCO and NuScenes~\cite{caesar2020nuscenes} datasets. 
We see that $r'$ is shifted towards large objects despite the actual ratio of those objects being relatively small. This forces the training to focus more on large objects, which benefits performance across all sizes. This raises the question of the ideal ratios for the distribution of object sizes when building a dataset, and likely this will depend on the target objects and their complexity at different sizes. Thus each dataset might have a different optimal weighting function.

\begin{table}
\centering
\resizebox{0.8\columnwidth}{!}{%
\begin{tabular}{|c|cc|cc|cc|}
\hline
         & \multicolumn{2}{c|}{Small} & \multicolumn{2}{c|}{Medium} & \multicolumn{2}{c|}{Large} \\
Dataset & \multicolumn{1}{l}{$r$} & \multicolumn{1}{l|}{$r'$} & \multicolumn{1}{l}{$r$} & \multicolumn{1}{l|}{$r'$} & \multicolumn{1}{l}{$r$} & \multicolumn{1}{l|}{$r'$} \\ \hline
COCO     & 0.28         & 0.13        & 0.44         & 0.33         & 0.29         & 0.54        \\
NuScenes & 0.39         & 0.15        & 0.46         & 0.38         & 0.15         & 0.47        \\ \hline
\end{tabular}%
}
\caption{Comparison of $r_{size}$ and $r'_{size}$ across all 
sizes on COCO and NuScenes, note how the weighted ratios $r'$ are shifted towards large objects compared to the normal ratios of objects.}
\label{tab:bars}
\end{table}

\begin{figure}
    \centering
    \includegraphics[width=.9\linewidth]{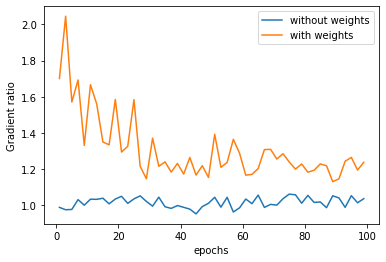}
\vspace{-.5em}

    \caption{Evolution of the ratio of the sum of gradient amplitudes $r_{grad}(\theta)$ over the first 100 epochs on COCO for YOLO v5. We see that the weighting term makes the impact of large object greater than small objects, resulting in an overall increase in performances. }
    \label{fig:grad}
\end{figure}

\begin{figure}
    \centering
    \includegraphics[width=.9\linewidth]{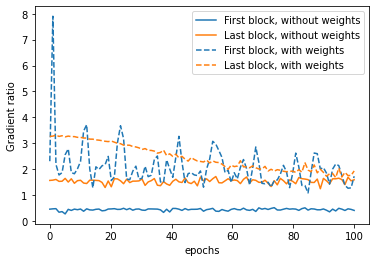}
\vspace{-.5em}

    \caption{Evolution of $r_{grad}(\theta_{block})$ restricted to the first and last BottleNeckCSP blocks for YOLO v5 over the first 100 epochs on COCO. We see that the both layers (especially the first one) are affected by the weighting function. }
    \label{fig:grad_blocks}
\end{figure}

\subsection{The effect of the weight term on the training}

In order  to gain more insight on the effect of the weighting term on the training, we need to quantify the importance of each sample during training. The authors of~\cite{vodrahalli2018all}  argue that the sum of the gradient magnitude of the loss can be a good measure of this. 
In fact, the evolution of the parameters of the model $\theta$   during training  is proportional to the magnitude of gradient of the loss  w.r.t the model parameters    $\left \|  \sum_{i \in Batch}^{} \nabla_{\theta} L_{i, \theta}\right \|$.
Since these gradients live in a high dimensional space, any two gradient vectors associated to two inputs are likely orthogonal. Therefore the triangular inequality
\begin{equation}
    \left \|\sum_{i \in Batch}^{}  \nabla_{\theta} L_{i, \theta} \right \| \leq \sum_{i \in Batch}^{} \left \| \nabla_{\theta} L_{i, \theta} \right \|
\end{equation} 
can be used as a tight estimate of the weights update.
Thus, we can consider $\left \| \nabla_{\theta} L_{i, \theta} \right \|$ as a measure of the impact of each object on the learned features and we can regroup these quantities by object size to see the effect of each object size on the learning procedure.
We computed the ratio of the sum of gradient magnitudes of large objects over those of small objects
\begin{equation}
    r_{grad}(\theta) = \frac{\sum_{i \in \Omega_{Large}}^{} \left \| \nabla_{\theta} L_{i, \theta} \right \|}{\sum_{i \in \Omega_{Small}}^{} \left \| \nabla_{\theta} L_{i, \theta} \right \|},
\end{equation}
where $\Omega_{Large}$ is the set of large objects, $\Omega_{Small}$ is the set of small objects and $L_{i,\theta}$ is the training loss term evaluated and the input $i$ (before the reduction over the image and the entire batch).

Figure~\ref{fig:grad} shows the evolution of this ratio along 100 epochs for YOLO v5 on COCO, using or not using the proposed weighting term. We can see that, without the weighting term, small and large object have an comparable contribution on the model parameters. This translates to $r_{grad}(\theta)$ oscillating around $1$. In contrast, using the weighting increases the impact of larger objects. This is shown by the value of $r_{grad}(\theta)$  starting high (at about $1.8$) at the beginning of the training  and stays larger than $1$ as the training continues.

To further investigate this effect, we studied this behavior at different levels of the network. The YOLO v5 architecture is based on 7 BottleNeckCSP blocks: two of them form the backbone and the others are the main component of the model's neck (the PANet part). We restrict the analysis to the parameters of the first or last BottleNeckCSP blocks and define
\begin{equation}
    r_{grad}(\theta_{block}) = \frac{\sum_{i \in \Omega_{Large}}^{} \left \| \nabla_{\theta_{block}} L_{i, \theta_{block}} \right \|}{\sum_{i \in \Omega_{Small}}^{} \left \| \nabla_{\theta_{block}} L_{i, \theta_{block}} \right \|},
\end{equation}
where $\theta_{block}$ is the set of parameters of a given BottleNeckCSP block of the model. 
Figure~\ref{fig:grad_blocks} shows the evolution of  $r_{grad}(\theta_{block})$ for the parameters of the first or last BottleNeckCSP blocks. 

This provides insights about the effects on low-level and high-level features. We see that the first block is particularly impacted when the weighting function is used, with the ratio increasing up to 16 folds at the beginning  of the training and stabilizing at a 4-fold increase later on. For the last layer, we still observe an increase of $r_{grad}$, but less important. 
This suggests that focusing the training on large objects impacts mostly the low-level features and does so during all the training. One can argue that these generic low-level features 
are more distinguishable on large objects than on small ones. 


These findings  shed some light on how the re-weighting  affects the training, suggesting that low level features are benefiting the most from large objects. 
In addition, one can argue that the shift of focus towards large objects is related to the  overall performance improvement as this  is observed  since the first training epochs (this will be discussed in the next section).


\section{Experiments}
To corroborate the impact of the proposed weighting scheme we compare the performance of several object detectors: YOLO V5, InternImage, DETR~\cite{carion2020end} and Mask R-CNN~\cite{he2017mask} on the COCO and nuScenes datasets, with and without the weight term.
We trained these models on both datasets on two NVIDIA RTX 2080 Ti for 35 epochs each with a batch size of 16. We used a warm-up of 5 epochs for InternImage-T. We used the Adam optimizer~\cite{kingma2017adam}
with a Cosine Annealing learning rate starting from a max value of 0.01 for YOLO v5 and Mask R-CNN and 0.1 for InternImage-T and DETR. The minimum IoU for validating a detection is fixed to 0.5 and a confidence threshold of 0.001 for COCO and 0.05 for nuScenes. As for data augmentation, we kept the same pipeline defined for each method on their respective papers. 

The results of these experiments, in terms of mAP and mAR scores are shown in Table~\ref{tab:scores}.
We see that all models exhibit  a significant performance improvement across all object sizes when using the proposed weighting scheme. For instance, InternImage-T with the proposed changes reaches 51.2\% mAP, while the original had 47.2\% mAP, which is a 4 p.p. gain. Our base results reproduce the results of InternImage's authors, and their paper shows that InternImage-B, which has more than double the number of parameters than InternImage-T, only reaches 48.8\% under similar training. We couldn't train with our modifications their biggest model InternImage-XL, which is the state of the art at the time of writing, as it requires expensive training resources. It is likely training such a model would define a new state of the art.  While the results are shown here on four different CNN object detectors, the proposed weighing scheme is quite simple and can be applied easily to other object detection models.

\begin{figure}

    \begin{tikzpicture}[
 image/.style = {text width=0.12\textwidth, 
                 inner sep=0pt, outer sep=0pt},
node distance =0mm and 8mm
                        ] 

\node[image] (frame4)
{\includegraphics[trim={50 50 50 50}, clip ,height=1.22\linewidth,width=1.22\linewidth]{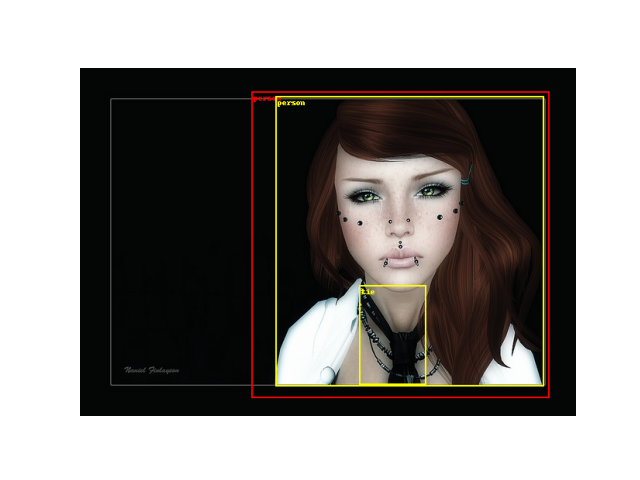}};
\node [image,right=of frame4] (frame5) 
    {\includegraphics[trim={50 50 50 50}, clip, height=1.22\linewidth,width=1.22\linewidth]{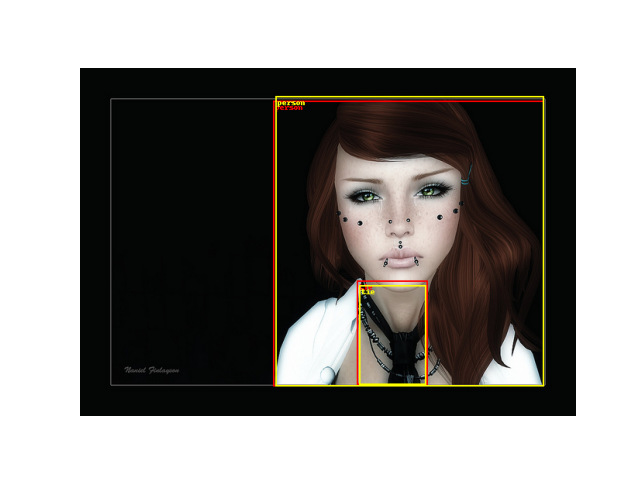}};
\node [image,right=of frame5] (frame6) 
    {\includegraphics[trim={50 50 50 50}, clip, height=1.22\linewidth,width=1.22\linewidth]{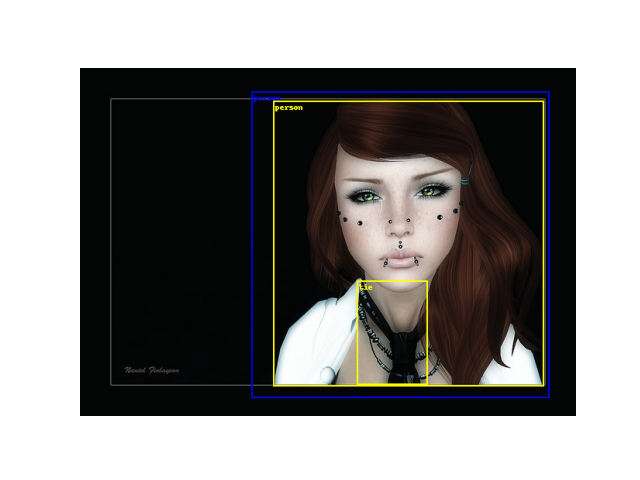}};

\node[image,below=of frame4] (frame7)
{\includegraphics[trim={50 50 50 50}, clip, height=1.22\linewidth,width=1.22\linewidth]{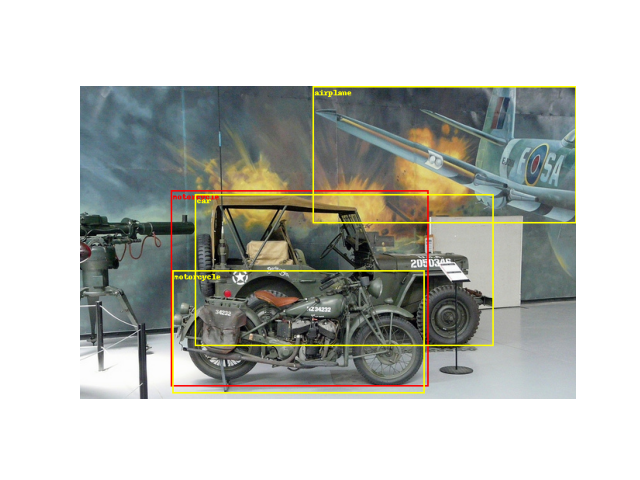}};
\node [image,right=of frame7] (frame8) 
    {\includegraphics[ trim={50 50 50 50}, clip, height=1.22\linewidth,width=1.22\linewidth]{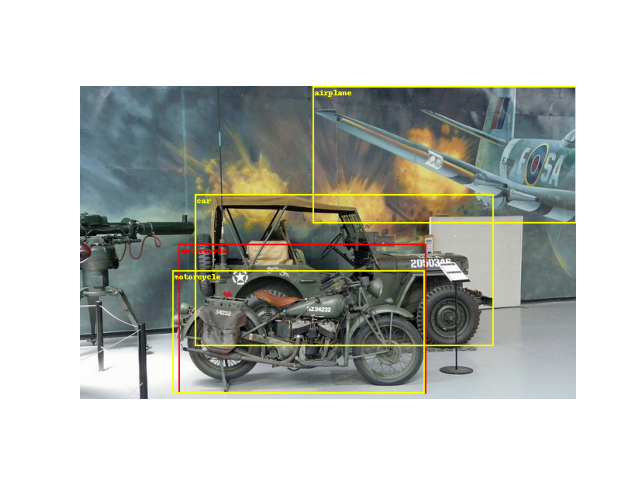}};
\node [image,right=of frame8] (frame9) 
    {\includegraphics[trim={50 50 50 50}, clip, height=1.22\linewidth,width=1.22\linewidth]{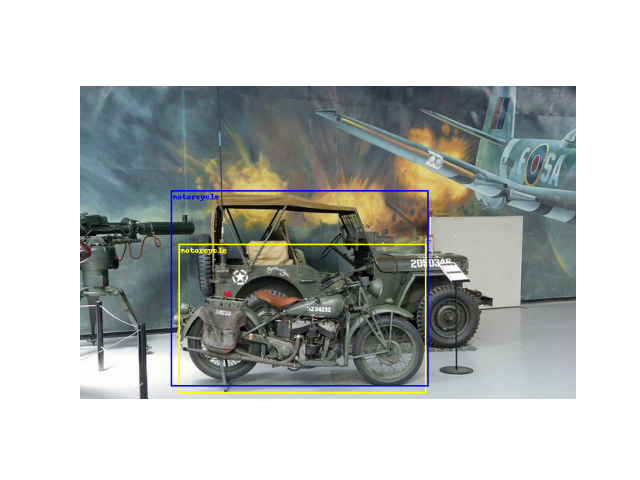}};

\node[image,below=of frame7] (frame10)
{\includegraphics[trim={50 50 50 50}, clip, height=1.22\linewidth,width=1.22\linewidth]{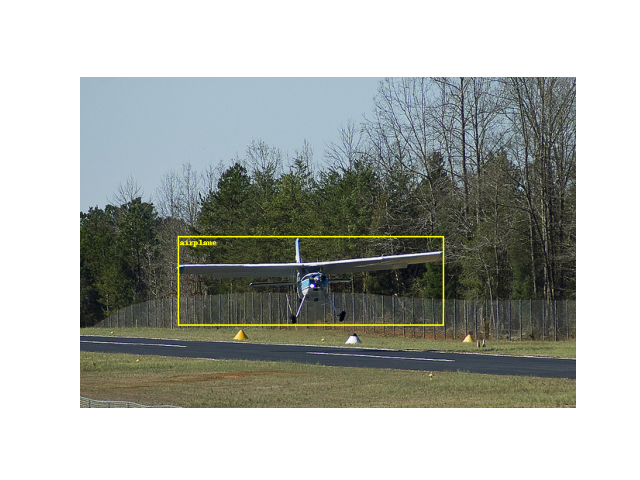}};
\node [image,right=of frame10] (frame11) 
    {\includegraphics[trim={50 50 50 50}, clip, height=1.22\linewidth,width=1.22\linewidth]{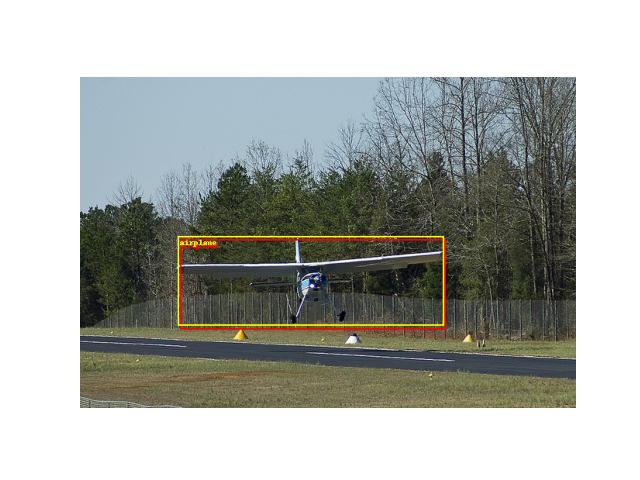}};
\node [image,right=of frame11] (frame12) 
    {\includegraphics[trim={50 50 50 50}, clip, height=1.22\linewidth,width=1.22\linewidth]{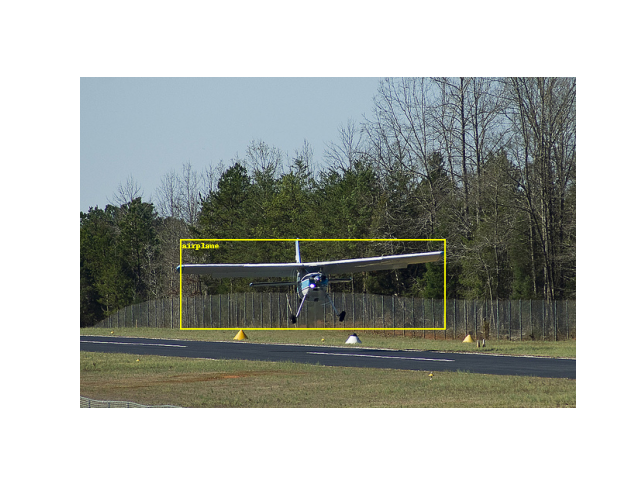}};

\end{tikzpicture}
    \caption{Qualitative display of some of the improvement made by adding the sample weight term. Columns from left to right: Without weights (Red: Prediction, Yellow: GT), With weights (Same Colors), Comparison (Blue: Without weights, Yellow: With weights)}
    \label{fig:qltvexmpls}
\end{figure}

A qualitative comparison is also shown in Figure~\ref{fig:qltvexmpls}. The selected examples show that the proposed modification allows the model to detect some objects that were otherwise undetected. For example, on the first and third rows, a tie and a plane are detected, respectively, only on the model with our modification. Bounding box predictions are also improved, as can be seen for example on the first and second row, where objects detected by both models have a more precise bounding box on the second column.

\begin{figure}
    \centering
    \includegraphics[trim={0 20 0 30}, clip, width=\linewidth]{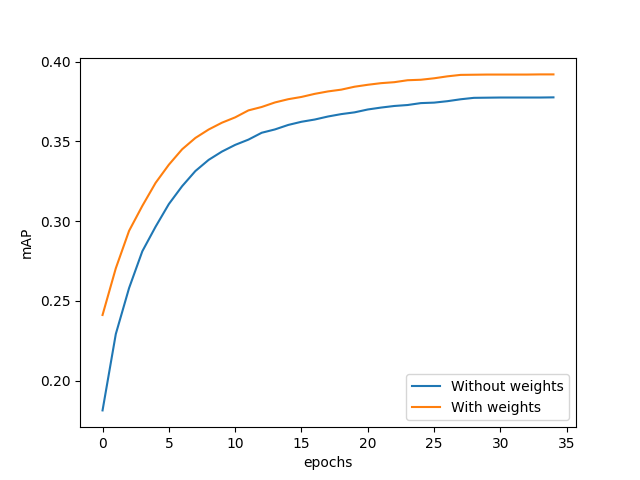}

    \caption{The addition of a weighting term to the detection and localization loss improves the validation scores on COCO val 2017 since the beginning of the training. }
    \label{fig:mapevo}
\end{figure}

\begin{table*}
\centering
\captionsetup{justification=centering}
\resizebox{1.8\columnwidth}{!}{
\begin{tabular}{|lcccccccccl|}
\hline
\multicolumn{3}{|c}{} &
  \multicolumn{2}{c}{Small} &
  \multicolumn{2}{c}{Medium} &
  \multicolumn{2}{c}{Large} &
  \multicolumn{2}{c|}{All} \\ \hline
Model &
  \multicolumn{1}{l}{\#Params} &
  \multicolumn{1}{l}{Weighted loss} &
  \multicolumn{1}{l}{mAP} &
  \multicolumn{1}{l}{mAR} &
  \multicolumn{1}{l}{mAP} &
  \multicolumn{1}{l}{mAR} &
  \multicolumn{1}{l}{mAP} &
  \multicolumn{1}{l}{mAR} &
  \multicolumn{1}{l}{mAP} &
  mAR \\ \hline
  \multicolumn{1}{|c}{{Mask R-CNN ResNet-50 FPN}} &
  {44M} &
  No &
  0.223 &
  0.386 &
  0.485 &
  0.621 &
  0.513 &
  0.697 &
  0.364 &
  0.512 \\
\multicolumn{1}{|c}{} &
   &
  Yes &
  \textbf{0.248} &
  \textbf{0.403} &
  \textbf{0.518} &
  \textbf{0.641} &
  \textbf{0.560} &
  \textbf{0.724} &
  \textbf{0.393} &
  \textbf{0.545} \\ \hline

\multicolumn{1}{|c}{{YOLO v5}} &
  {64.4M} &
  No &
  0.297 &
  0.456 &
  0.520 &
  0.640 &
  0.617 &
  0.745 &
  0.372 &
  0.599 \\
\multicolumn{1}{|c}{} &
   &
  Yes &
  \textbf{0.315} &
  \textbf{0.462} &
  \textbf{0.549} &
  \textbf{0.702} &
  \textbf{0.654} &
  \textbf{0.781} &
  \textbf{0.398} &
  \textbf{0.623} \\ \hline
  \multicolumn{1}{|c}{{DETR ResNet-50}} &
  {41M} &
  No &
  0.312 &
  0.459 &
  \textbf{0.531} &
  0.675 &
  0.628 &
  0.776 &
  0.420 &
  0.553 \\
\multicolumn{1}{|c}{} &
   &
  Yes &
  \textbf{0.331} &
  \textbf{0.460} &
  0.529 &
  \textbf{0.697} &
  \textbf{0.659} &
  \textbf{0.802} &
  \textbf{0.440} &
  \textbf{0.570} \\ \hline

\multicolumn{1}{|c}{{InternImage-T}} &
  {49M} &
  No &
  0.309 &
  0.464 &
  0.538 &
  0.718 &
  0.674 &
  0.833 &
  0.472 &
  0.66 \\
 &
   &
  Yes &
  \textbf{0.332} &
  \textbf{0.487} &
  \textbf{0.556} &
  \textbf{0.736} &
  \textbf{0.714} &
  \textbf{0.857} &
  \textbf{0.512} &
  \textbf{0.679} \\ \hline
\end{tabular}%
}
\caption{Models performances on COCO val 2017. We can see from the results that the introduction of the sampling weight term improved the models scores across all sizes.}
\label{tab:scores}
\end{table*}

\begin{table*}
\centering
\captionsetup{justification=centering}
\resizebox{1.8\columnwidth}{!}{
\begin{tabular}{|lcccccccccl|}
\hline
\multicolumn{3}{|c}{} &
  \multicolumn{2}{c}{Small} &
  \multicolumn{2}{c}{Medium} &
  \multicolumn{2}{c}{Large} &
  \multicolumn{2}{c|}{All} \\ \hline
Model &
  \multicolumn{1}{l}{\#Params} &
  \multicolumn{1}{l}{Weighted loss} &
  \multicolumn{1}{l}{mAP} &
  \multicolumn{1}{l}{mAR} &
  \multicolumn{1}{l}{mAP} &
  \multicolumn{1}{l}{mAR} &
  \multicolumn{1}{l}{mAP} &
  \multicolumn{1}{l}{mAR} &
  \multicolumn{1}{l}{mAP} &
  mAR \\ \hline
{InternImage-T} &
  {49M} &
  No &
  0.551 &
  0.634 &
  \textbf{0.573} &
  \textbf{0.710} &
  0.652 &
  0.738 &
  0.648 &
  0.711 \\
 &
   &
  Yes &
  \textbf{0.562} &
  \textbf{0.661} &
  0.570 &
  0.708 &
  \textbf{0.679} &
  \textbf{0.762} &
  \textbf{0.659} &
  \textbf{0.724} \\ \hline
\end{tabular}%
}
\caption{InternImage-T performances on NuScenes. These results show that the benefits of the weighting term are reproducible on different datasets. The amount of progress may depend on the ratios of small and large objects.}
\label{tab:scores2}
\end{table*}

We also validate the improvement on another dataset: NuScenes. We used InternImage as model and compared its performance with and without the weight term. The results are shown in Table~\ref{tab:scores2}.
We observe that we still have a slight improvement in the scores with the weighted loss. 
The evolution of the overall mAP w.r.t the number of epochs, shown in Figure~\ref{fig:mapevo}, proves also that the model benefits from focusing on  big objects since the start of the training as the performance is consistently better along the training procedure. We can see that from the first epochs, our weighting policy yields an improvement of nearly 3 p.p. on average. This emphasizes the intuition that the increased presence of large objects helps steer training in a better direction and avoids worse local minima. This also indicates that the effect of future improvements to object weighting could be seen early in training.

\begin{table*}
\centering
\resizebox{1.8\columnwidth}{!}{
\begin{tabular}{|ccccccccccc|}
\hline
$L_{classif}$ & $L_{detection}$ & $mAP_{small}$  & $mAP_{medium}$ & $mAP_{large}$  & $mAP_{all}$  & $MAE_{small}$  & $MAE_{medium}$ & $MAE_{large}$  & $MAE_{all}$ & $AP@50_{all}$    \\ \hline
-             & -               & 0.297          & 0.540          & 0.674          & 0.372    & 9.843          & 12.427         & 14.824         & 12.523  & 0.572    \\
\checkmark    & -               & 0.290          & 0.542          & 0.680          & 0.371      & 8.245          & 11.346         & 11.293         & 10.576   & 0.584   \\
-             & \checkmark      & 0.315          & 0.551          & 0.686          & 0.384       & 0.315          & 0.551          & 0.686          & 0.384   & 0.587    \\
\checkmark    & \checkmark      & \textbf{0.332} & \textbf{0.556} & \textbf{0.714} & \textbf{0.398}  & \textbf{5.645} & \textbf{9.587} & \textbf{8.386} & \textbf{8.231} & \textbf{0.615}  \\ \hline
\end{tabular}%
}
\caption{Ablation study on the introduction of the weight term in classification and detection loss term and its effect on the mAP@50:95, the Mean Absolute Error (Pixels) on the bounding box center and the Average Precision on IoU=0.5 on COCO val 2017 dataset.}
\label{tab:abl}
\end{table*}

\section{Ablation studies and discussion}

\begin{table}
\centering
\captionsetup{justification=centering}
\resizebox{0.6\columnwidth}{!}{%
\begin{tabular}{|c|c|c|}
\hline
Sample weight function              & mAP          & mAR          \\ \hline
$f(w \times h) = 1$        & 0.372        & 0.599        \\
$f(w \times h) = w \times h$        & 0.217        & 0.412        \\
$f(w \times h) = \sqrt{w \times h}$ & 0.243        & 0.459        \\
$f(w \times h) = log(w \times h)$   & \textbf{0.398}        & \textbf{0.623}        \\ \hline
\end{tabular}%
}
\caption{Results of different sample weights functions on COCO 2017 val using YOLO v5.}
\label{tab:abl2}
\end{table}

\begin{figure}[t]
    \centering
    \includegraphics[trim={0 5 0 30}, clip, width=\linewidth]{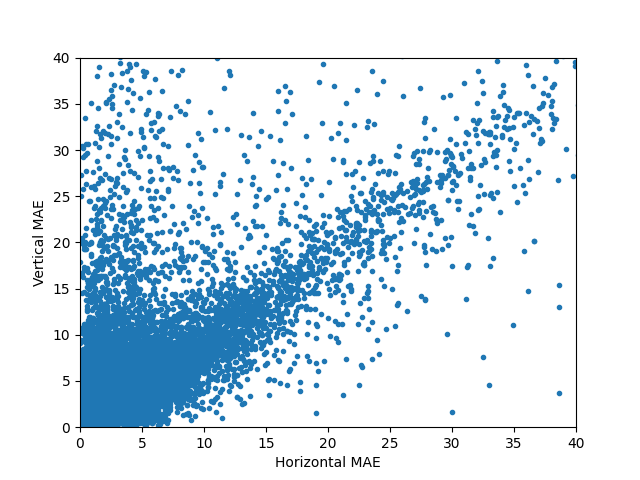}

    \caption{The horizontal and vertical MAEs are highly correlated for YOLO v5 on COCO, with a correlation coefficient of 0.7710 }
    \label{fig:mae_corr}
\end{figure}
\subsection{Impact on the terms of the loss}
\label{sub:ab}
To further investigate the impact of weighting strategies in the YOLO v5 loss function, we conducted an ablation study on the COCO dataset. Given the total loss function of the model  \eqref{eq:loss}, we vary the use of the weighting function for both $L_{classif}$ and $L_{detection}$. More specifically, we explored four scenarios: no weight terms, weight terms applied to the classification term only, weight terms applied to the detection term only and the weight term applied to all the loss terms.

Our analysis focuses on evaluating the Mean Average Precision (MAP@50:95) as a general metric score and the error on the bounding box center as a localization metric score. Table~\ref{tab:abl} shows the impact of each combination on the mAP for various sizes of objects. As the mAP is impacted by both the localization errors and the capacity of the network to detect the objects and correctly classify them, we complement the results the Mean Absolute Error (MAE: average L1 distance of the predicted bounding box center compared to the ground truth center). The MAE was estimated only on the horizontal component. We can justify this by the high correlation between horizontal and vertical MAEs (see Figure~\ref{fig:mae_corr}).
In order to reduce the impact of the capacity of the network to detect objects, these results were computed on the set of objects correctly detected (correct class and IOU $>$ 0.5). Lastly as AP@50 is less sensitive to localization errors, we display the corresponding results across all objects.

The results show that when adding only the weighting scheme to the classification term, the mAP regresses slightly, in particular for small objects, despite improved AP50 and MAE. The exact interpretation of this phenomenon is unclear. However, when the changed term is the detection term, mAP, MAE and AP50 are improved. The MAE gain is more important relatively for large objects (30\%), indicating better localization. Lastly, having the weighting scheme on both loss terms gives the best performance on all metrics. Proportionally compared to the initial results, the highest gain is seen on small targets, as it sees a 12 p.p. increase in mAP (against 3 p.p. for medium and 6 p.p. for large objects) and a 43\% decrease in MAE (against 23\% for medium and 36\% for large objects).

This suggests that a holistic approach that considers both classification and detection, with the weight terms appropriately assigned, is crucial for achieving the best results in terms of mAP score and bounding box center error.

\subsection{On the choice of $log(w \times h)$}

As discussed above, the main idea behind the choice of $log(w \times h)$ is to increase the contribution of large objects in the learning of the network features. 
We tested  other functions of $w \times h$ and compared them to the proposed function. Table~\ref{tab:abl2} evaluates some  sample weighting functions for YOLO v5 on the COCO dataset. We kept on the idea that this function should depend on the areas of the objects and changed only the type of function (linear, logarithmic, square root). 
Although $log(w \times h)$ yields the best results in this table, we believe that additional research and experimentation is required in this direction in order to identify better functions or to prove that the chosen weight function is the optimal choice for better performances.
\subsection{Impact of the dataset}
The performance gain was demonstrated on two datasets: COCO and NuScenes. While the performance gain on these two datasets is far from negligible, there is no guarantee that similar gains can be obtained on other datasets. In fact the weighting scheme comes down to artificially increase the proportion of larger objects in the dataset, and thus if the dataset already had optimal proportions, the weighting wouldn't increase the performance. The conclusions from this research though is that when building a dataset, it is important to have a significant proportion of large objects, and if not, compensate with a weighting factor.
One aspect impacting weighting needs is the difficulty of the detection of each object's size. For COCO and NuScenes, the detection scores for small objects are lower than for large objects. As small objects are harder to detect they tend to have stronger errors in the loss, and thus higher gradients. The weighting scheme can be seen as a correction factor to this behavior. 

\section{Conclusion}
In this paper, we have shown that the presence of large objects in the training dataset helps to learn features that yield better performance also on small and medium objects. 
We then proposed a simple loss reweighting scheme that leads to improved performance of object detectors.
Our findings underscore the importance of considering large objects and demonstrate the potential of incorporating a weighted loss term in enhancing overall object detection performance.
Through experiments and ablation studies, we validated the effectiveness of our proposed approach. We evaluated different models and datasets, consistently observing improvements in detection scores across all sizes.

Future research in this area could investigate novel strategies that explicitly consider the impact of large objects on detection accuracy across different scales.

\vspace{-1em}
\paragraph{Acknowledgments}

We acknowledge support from ANRT CIFRE Ph.D. scholarship n$^\circ$2020/0153 of the MESRI. This work was performed using HPC resources 
from GENCI–IDRIS (grant 2023-AD011011801R3) and from the “Mésocentre” computing center of CentraleSupélec and ENS 
Paris-Saclay supported by CNRS and Région Île-de-France (http://mesocentre.centralesupelec.fr/).

{\small
\bibliographystyle{ieee_fullname}
\bibliography{refs}
}
\end{document}